\documentclass[runningheads]{llncs}

\usepackage{times}
\usepackage{epsfig}
\usepackage{graphicx}
\usepackage{amsmath}
\usepackage{amssymb}

\DeclareMathOperator*{\argmax}{arg\,max}

\DeclareMathOperator*{\Card}{Card}

\usepackage{mathrsfs}

\usepackage[utf8]{inputenc}

\usepackage[normalem]{ ulem }
\usepackage{soul}

\usepackage{algorithm}
\usepackage{algorithmic}

\usepackage{subfig}
\usepackage{xcolor}

\usepackage{authblk}
\usepackage{hyperref}

\newcommand{\Src}{S} 
\newcommand{\Tgt}{T} 

\title{Database Annotation with few Examples:\\ An Atlas-based Framework using Diffeomorphic Registration of 3D Trees}
\titlerunning{Atlas-based Vascular Tree Annotation}

\author{Pierre-Louis Antonsanti\inst{1,2} \and
Thomas Benseghir\inst{1} \and
Vincent Jugnon\inst{1} \and
Joan Glaunès\inst{2}}

\authorrunning{P. Antonsanti et al.}

\institute{GE Healthcare, Buc 78530, France \\ \email{\{pierrelouis.antonsanti,thomas.benseghir,vincent.jugnon\}@ge.com} \and
MAP5, Universite de Paris, Paris 75006, France \email{alexis.glaunes@parisdescartes.fr}}

\begin{document}

\maketitle

\begin{abstract}
    Automatic annotation of anatomical structures can help simplify workflow during interventions in numerous clinical applications but usually involves a large amount of annotated data.
    The complexity of the labeling task, together with the lack of representative data, slows down the development of robust solutions.
    In this paper, we propose a solution requiring very few annotated cases to label 3D pelvic arterial trees of patients with benign prostatic hyperplasia.
    We take advantage of Large Deformation Diffeomorphic Metric Mapping (LDDMM) to perform registration based on meaningful deformations from which we build an atlas.
    Branch pairing is then computed from the atlas to new cases using optimal transport to ensure one-to-one correspondence during the labeling process.
    To tackle topological variations in the tree, which usually degrades the performance of atlas-based techniques, we propose a simple bottom-up label assignment adapted to the pelvic anatomy.
    The proposed method achieves 97.6\% labeling precision with only 5 cases for training, while in comparison learning-based methods only reach 82.2\% on such small training sets.

\end{abstract}

\section{Introduction}

The automatic annotation of tree-like structures has many clinical applications,
from workflow simplification in cardiac disease diagnosis (\cite{Akinyemi09}, \cite{Cao17}, \cite{Ezquerra1998},  \cite{Wu18}) to intervention planning in arterio-venous malformations (\cite{Bogunovic13}, \cite{Ghanavati14}, \cite{Robben16}, \cite{Wang17}) and lesion detection in pneumology (\cite{Feragen15}, \cite{VanGinneken08}, \cite{Lo11}). 
This task is particularly important in the context of interventional radiology, where minimally invasive procedures are performed by navigating small tools inside the patient's arteries under X-ray guidance. 
For example, benign prostatic hyperplasia symptoms can be reduced by embolizing arteries feeding the prostate to reduce its size \cite{Ray18}.

Identifying the correct arteries to treat during the procedure - along with their neighbours - is crucial for the safety of the patient and the effectiveness of the treatment.
However, this task is very challenging because of the arterial tree complexity and its topological changes induced by frequent anatomical variations.
Having a large representative annotated database is also a challenge, especially in medical imaging where the sensitivity of the data and the difficulty to annotate slow down the development of learning based techniques \cite{Lee2017medical}.
In this context, solutions to the tree labeling problem should ideally work from a limited number of annotated samples.

\subsubsection{Learning-based labeling}
With the prevalence of machine learning, many articles address the problem of automatic anatomical tree labeling by extracting features from the tree to feed a learning algorithm that predicts labels probabilities. 
Authors of \cite{VanGinneken08} and \cite{Akinyemi09} first proposed Gaussian mixtures models to learn from branch features (geometrical and topological) and predict the label probabilities. In \cite{Akinyemi09}, labels are assigned following clinical a priori defining a set of rules.
In \cite{Lo11}, k-Nearest-Neighbours predicts label probabilities that are used in a bottom-up assignment procedure, searching through all the existing branch relationships in the training data.
Since numerous features leads to high dimensional space, such techniques often show poor generalization capacity.

To reduce dimensionality, boosting algorithms and tree classifiers are used in \cite{Hoang11}, \cite{Matsuzaki14} and \cite{Robben16} to select the most discriminant features.
Assignment procedure is enriched by topological rules to improve coherence along the tree, instead of considering branches independently.
Label assignment was further refined in \cite{Ghanavati14} and \cite{Wang17} by adding a Markovian property - each branch label depending on its direct neighbours - with Markov chains parameters learned during training. Such assignment strongly depends on the database size and the task complexity in term of anatomical variability.

While previous articles had access to limited size databases (around 50 cases), authors of \cite{Wu18} trained a recurrent neural network preserving the topology of the coronary tree on a database composed of 436 annotated trees. The labels predictions rely on a multi-layer perceptron and a bidirectional tree-structural long short-term memory network. This interesting approach using deep learning for vessel classification is less common because it requires a lot more training data to be able to capture the anatomical variability.

\subsubsection{Atlas-based labeling}

Contrary to learning-based techniques, atlas-based ones can offer robust annotation even with few annotated cases (\cite{Wu18}).
An atlas is defined as a reference model that can be built from prior knowledge (\cite{Bogunovic13}, \cite{Cao17}) or from an available annotated database (\cite{Bulow06}, \cite{Ezquerra1998}, \cite{Feragen15}). Most of the atlas-based methods follow a four steps framework: the choice of the atlas, the registration onto the target, the estimation of the labels probabilities and finally the assignment.
In \cite{Ezquerra1998}, authors focused on the annotation procedure, relying on previously established atlas and a manual registration. 
Labeling is done through a branch-and-bound algorithm extending the best partial labeling with respect to a function designed to compare observations to the atlas. 

Later in \cite{Bulow06}, label probabilities are computed at the level of branches by a voting procedure, each point along a branch voting for a label.
This solution can adapt to missing branches but still does not guarantee anatomical consistency of the labeling along the tree and is not robust to topological variation with respect to the atlas.
To take this variability into account \cite{Bogunovic13} proposes to create one atlas per known topology in the anatomy of interest, an interesting approach if the number of anatomical variants is limited.
Similarly \cite{Cao17}, \cite{Feragen15}  use multi-atlas approaches, taking advantage of a distance to the atlas that quantifies topological differences. 
In \cite{Cao17}, the reference case is selected as the best example in a training set following a leave-one-in cross-validation design. 

\begin{figure}[t]
\centering
  \includegraphics[clip, width=1\textwidth]{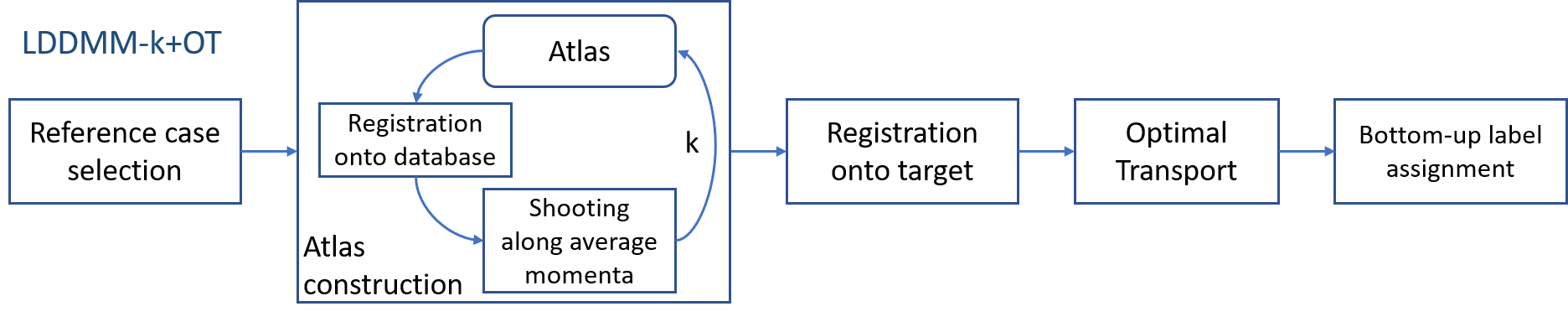}
\caption{\label{fig:schema} The proposed atlas-based vascular tree annotation pipeline.}
\end{figure}

While a lot of advanced methods have been proposed for medical structure registration \cite{Sotiras13}, few efforts have been made to use advanced deformation models in the context of vascular tree labeling \cite{Matl2017}.
In our work, we propose an atlas-based algorithm illustrated in Fig.~\ref{fig:schema} relying on state of the art LDDMM to build the atlas and estimate realistic deformations.
It is combined with an Optimal Transport based matching, to compute a relevant assignment between atlas and target branches.
In the end, to handle the topological changes between the atlas and the target, a bottom-up label assignment is performed to achieve optimal results for our pelvic vasculature labeling problem.


\section{Method}\label{sec:method}

Let a labeled vascular tree $\Tgt = \left( \left\{ (B_\alpha, l_\alpha) \right\}_\alpha , M \right)$ be a set of branches $B_\alpha$ labeled $l_\alpha \in \mathbb{N}$ with connections to other branches stored in an adjacency matrix $M$.
Two labeled branches $(B_\alpha, l_\alpha)$ and $(B_\beta, l_\beta)$ are connected if $M_{\alpha,\beta}=1$.
Each branch $B_\alpha$
is a polygonal curve composed of ordered points $B_\alpha[k] \in \mathbb{R}^3$ representing a vessel centerline. We also denote $\{q_i\}_i$ the unordered set of all points $B_\alpha[k]$ of the tree for all indices $\alpha$ and $k$.
\subsubsection{The Large Deformation Diffeomorphic Metric Mapping framework}

We compute the deformation $\varphi$ of a \textit{source} shape $\Src$  (in our case the atlas, a centerlines tree) onto a \textit{target} shape $\Tgt$ using LDDMM.
This state of the art framework, detailed in \cite{Younes10} (chapters 8 to 11), allows to analyze differences between shapes via the estimation of invertible deformations of the ambient space that act upon them. 
In practice, the diffeomorphism $\varphi$ is estimated by minimizing a cost function $\mathcal{J}(\varphi) = \mathcal{E}(\varphi)+\mathcal{A}(\varphi(\Src),\Tgt)$ where $\mathcal{E}$ is the deformation cost, and $\mathcal{A}$ is a data attachment term that penalizes mismatch between the deformed source $\varphi(\Src)$ and the target $\Tgt$. 
We chose the Normal Cycles model proposed in \cite{Roussillon16} as data attachment since it has shown good results at areas of high curvature and singular points as bifurcations or curves endpoints.

The deformation map $\varphi:\mathbb{R}^3\rightarrow\mathbb{R}^3$ is defined as the solution at time $t=1$ of a flow equation $\partial_t\phi(t,x)=v(t,\phi(t,x))$ with initial condition $\phi(0,x)=x$,
where $v(t,\cdot)$ are time varying vector fields assumed to belong to a Hilbert space $V$, ensuring regularity properties.
As shown in \cite{Miller06}, in a discrete setting, denoting $x_i, \: i \in [1,...,n_S]$ the discretization points of the source shape, one may derive optimality equations that must be satisfied by the trajectories $q_i(t)=\phi(t,x_i)$ when considering deformations that minimize the cost function. These equations take the following form: 
\begin{equation}\label{method:geodesics} 
\begin{cases}
   \displaystyle\dot{p}_i(t) = -\dfrac{1}{2}  \nabla_{q_i(t)}\left( \sum_{j=1}^{n_S}\sum_{l=1}^{n_S}\left\langle p_j(t)\;,\;K_V(q_j(t),q_l(t))p_l(t) \right\rangle \right) \\
    \displaystyle\dot{q}_i(t) = \sum_{j=1}^{n_S} K_V(q_i(t),q_j(t))p_j(t)
\end{cases}
\end{equation}
where $p_i(t)$ are auxiliary dual variables called momenta.
They correspond to geodesic equations with respect to a specific Riemannian metric, written in Hamiltonian form.
The deformations are then fully parametrized by the set of initial momenta $(p_i(0))_i \in \mathbb{R}^{3n_\Src}$.

To model deformations, we define the kernel to be a sum of Gaussian kernels $K_V(x,y) = \sum_s \exp\left(-\Vert x-y\Vert^{2}\;/\;(\sigma_0/s)^{2}\right)$, where $s \in [1,4,8,16]$ and $\sigma_0$ is half the size of our vascular trees bounding box.
This multi-scale approach was introduced by \cite{Risser11}. Normal Cycles also require to choose specific kernels; we take a sum of constant and linear kernels for the spherical part and a Gaussian kernel for the spatial part, at two scales $\sigma_0$ and $\sigma_0/4$, using the output $(p_i(0))_i$ of the optimization process at scale $\sigma_0$ as initialization of the optimization at scale $\sigma_0/4$.

\subsubsection{Building the atlas}\label{Atlas}

In the context of vascular tree labeling, the tree topology can be highly variable (in our database we have one topology every two cases). Trying to build atlases that take into account these variations is unrealistic. Yet we still want the automatic annotation to be robust to the choice of the reference case.

The LDDMM framework allows to derive methods for computing such atlases.
More precisely, the optimal deformations generated by LDDMM are fully parameterized by the initial momenta, and their representation in the Euclidean space $\mathbb{R}^{3n_S}$ allows to perform classical linear statistics. 
Following \cite{Vaillant04}, in order to build the atlas we select one available annotated case and compute its deformations onto a set of $N$ targets (the selected case included).
These registrations provide a collection of initial momenta $\{\left(p_i^k(0)\right)_i,k \in [1,...,N]\} $.
The reference case is selected using a leave-one-in
method as in \cite{Cao17} with respect to the labeling procedure.
The atlas is then obtained by shooting via geodesic equations to deform the selected case along the average of the initial momenta $ \overline{p_i(0)} = (1/N) \sum_k p_i^k(0)$.
This procedure does not require the targets to be annotated, and seems suited to build an atlas of the whole database.

In addition it can be iterated by replacing the selected case with the deformed one. 
We refer to \textbf{LDDMM-k} for the k-th iteration of the atlas construction. 
Consequently LDDMM-0 refers to the case of using the reference case directly as an atlas. 
As illustrated in Fig.~\ref{fig:mean_shoot} the atlas converges along the iterations to an average position representative of the set of targets.
Through this spatial normalization, we limit the sensitivity to the choice of the initial case as atlas.
To illustrate the impact of LDDMM-k on the labeling, we chose a simple assignment procedure described in \cite{Bulow06}: each point of the target tree votes for the label of its closest point in the labeled atlas.
Then, we compute the vote per branch $B_\beta^\Tgt$ in the target:
$\pi(B_\beta^\Tgt,l)=\Card( \{ q_i^\Tgt\in B_\beta^\Tgt,\; \hat{l}(q_i^\Tgt)=l\})\;/\;\Card(B_\beta^\Tgt)$.
This label probability estimation does not guarantee anatomical consistency and deeply relies on the registration. 
Additionally, each point vote is independent from the others, allowing to characterize the quality of the registration from a labeling point of view.

\subsubsection{Optimal Transport for a better assignment}

During the LDDMM atlas construction and registration, each tree is seen as one shape. Consequently, there is no assumption over branch matching and topological changes.
In order to provide a relevant label assignment that takes the mutual information into account we propose to use Optimal Transport. 
It is convenient to compute the optimal one-to-one assignment between branches of the deformed source and the target with respect to a given distance. 

Based on the work of \cite{Feydy19}, each branch is re-sampled with 20 points and the distance matrix $D$ between each branch $B_\alpha^{\varphi(\Src)}$ and $B_\beta^\Tgt$ is given by: $ D_{\alpha,\beta} = \Vert B_{\alpha}^{\varphi(\Src)} - B_{\beta}^{\Tgt} \Vert_{\mathbb{R}^{3d}}$.
We tried different numbers of points $d$ per branch ranging from 20 to 500 with no significant impact over the matching results.  
Considering that our problem is of limited size (17 branches per tree), a simple Kuhn-Munkres algorithm (also called Hungarian algorithm) was used to compute the assignment solution. 
It consists in finding minimum weight matching in bipartite graphs by minimizing the function $\sum_{\alpha,\beta} D_{\alpha,\beta}.X_{\alpha,\beta}$ with $\left(X_{\alpha,\beta}\right)\in \{0,1\}^{17 \cdot 17}$ the output boolean matrix with $1$ if the branches $B_\alpha^{\varphi(\Src)}$ is assigned to $B_\beta^\Tgt$, $0$ otherwise.
To be consistent with the alternative simple label probability estimation of our LDDMM-k pipeline, we similarly define here $\pi(B_\beta^\Tgt,l_\alpha^{\Src}):=X_{\alpha,\beta}$, although these "probabilities" are always $0$ or $1$ in this case.
This assignment process is complementary to
the LDDMM-k process since it focuses on assignment between branches while LDDMM-k focuses on the atlas construction and the registration. We will call this pipeline \textbf{LDDMM-k+OT}.
We will also experiment the Optimal Transport assignment without any registration (i.e. taking $\varphi = id$), which we denote \textbf{OT}.

\subsubsection{Label assignment post-processing}

A first label assignment procedure directly takes the highest label probability for each branch: $ \hat{l}(B_\alpha^\Tgt) = \argmax_l(\pi(B_\alpha^\Tgt,l))$, $\pi$ being the output of the OT procedure or the voting probabilities estimation.
This \textit{direct assignment} is not based on a priori knowledge and directly reflects the performance of the prediction methods. 

In practice, the vessels are labeled by the expert accordingly to the anatomy or area they irrigate \cite{Assis15}.
In the application to pelvic vascular tree, when two branches of different labels share a parent, this parent is called \textit{"Common Artery"}. 
This is the only clinical a priori we introduce in the method.
To limit the effect of the topological variations between the atlas and the target, we propose a \textit{bottom-up assignment} procedure:
first for all $B_\alpha^\Tgt$ leaf of $\Tgt$, $\hat{l}(B_\alpha^\Tgt) = \argmax_l(\pi(B_\alpha^\Tgt,l))$
then recursively every parent branch $B_\alpha^\Tgt$ 
is assigned a label with the rule:
$$\hat{l}(B_\alpha^\Tgt) = \left\{
    \begin{array}{ll}
        l\quad \mbox{  if  } \hat{l}(B_\beta^\Tgt)=l \quad\text{for every branch }B_\beta^\Tgt\text{ child of }B_\alpha^\Tgt \\
        0\quad (i.e. "Common \; Artery") \mbox{,  otherwise.}
    \end{array}
\right.
$$
This recursive assignment procedure, although specific to this anatomy, is quite adaptable. 
In fact, in most of the structured tree-shaped anatomies (coronary \cite{Akinyemi09}, \cite{Cao17}, airway tree \cite{Lo11}, pelvic \cite{Assis15}) the branches names are also conditioned by the leaves labels. 
When two arteries of different labels share a common parent, this parent is either unnamed (as in our application), or named by a convention provided by the experts. 
The latter situation corresponds to additional conditions (as in \cite{Lo11}) during the assignment. 

\section{Results}
 
\begin{figure}[t]
\minipage[t]{0.25\textwidth}
\centering
  \includegraphics[trim={5cm 2cm 5cm 2cm}, clip, width=1\textwidth]{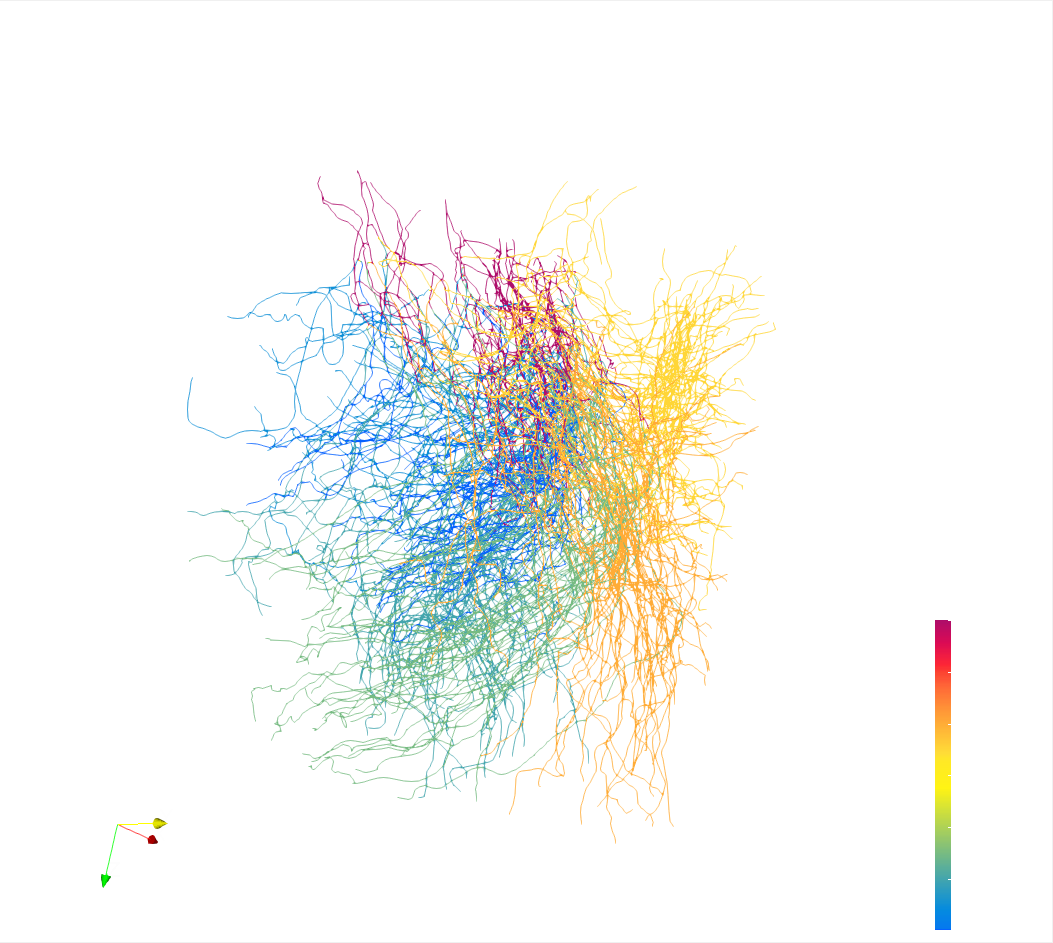}
  \caption*{(a)}\label{fig:RawTrees}
\endminipage\hfill
\centering
\minipage[t]{0.25\textwidth}
  \includegraphics[trim={5cm 2cm 5cm 2cm}, clip, width=1\textwidth]{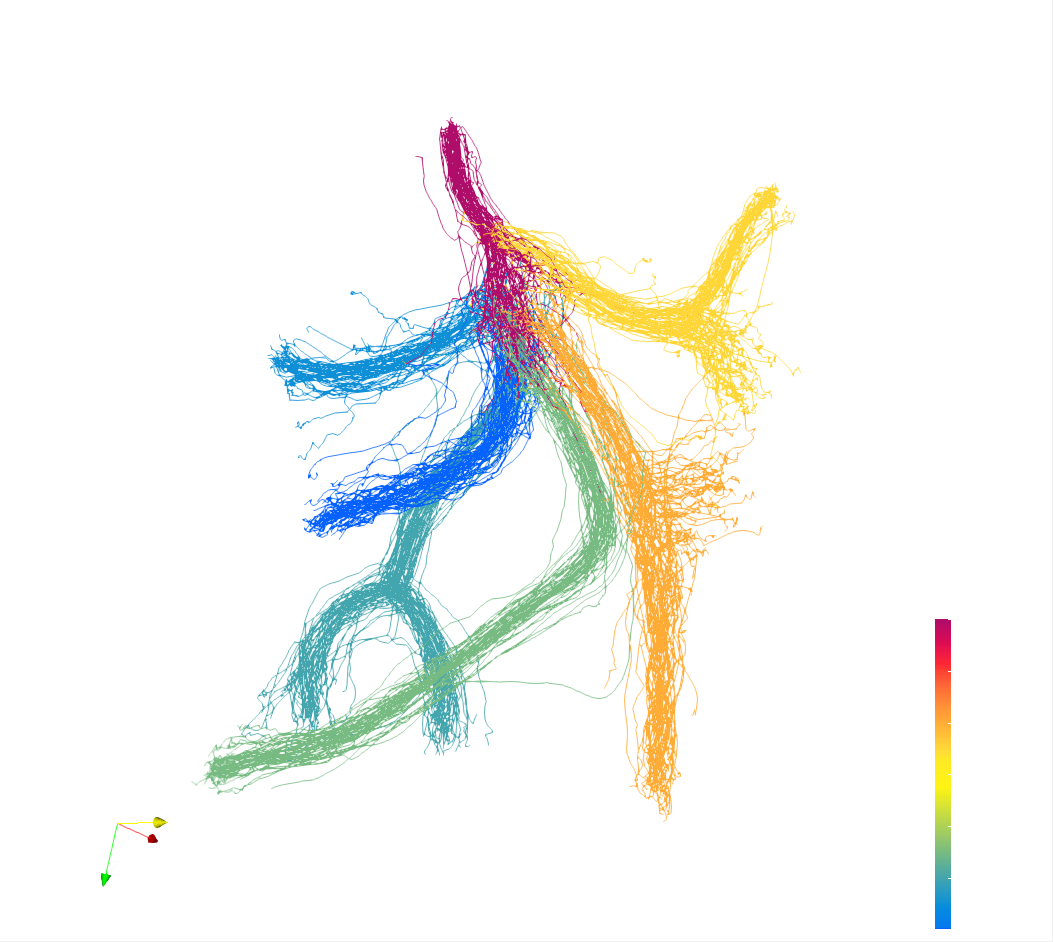}
  \captionsetup{labelformat=empty}
  \caption*{(b)}\label{fig:average_shooting}
\endminipage\hfill
\minipage[t]{0.5\textwidth}%
  \includegraphics[clip, width=1\textwidth]{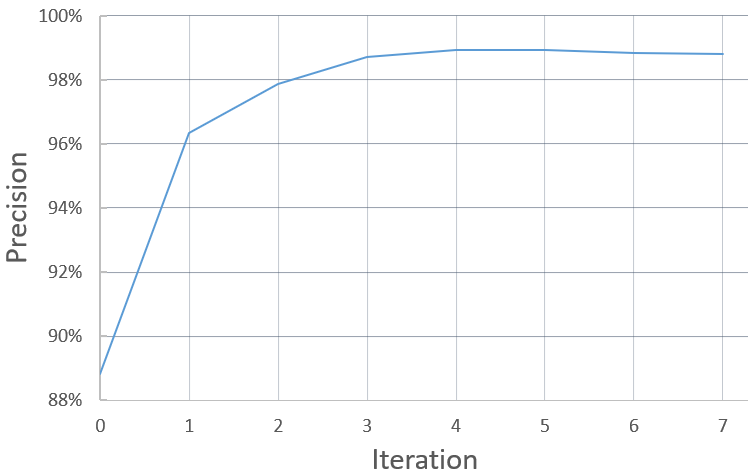}\\
  \captionsetup{labelformat=empty}
  \caption*{(c)}\label{fig:template_iteration}
\endminipage
\caption{\label{fig:mean_shoot} Building atlases via LDDMM registration. The colors represent the ground truth labels. (a) The initial trees; (b) Reference atlases at LDDMM-1 (c) Precision of one atlas over LDDMM-k iterations.}
\end{figure}

We conducted experiments on a dataset of 50 pelvic vascular trees corresponding to 43 different patients, some trees being the left and right vasculature of a single patient.
The centerlines composing the vascular tree are constructed from 3D volumes (injected Cone Beam Computed Tomography).
While the entire vascular tree is composed of up to 300 different branches, we manually extracted a simplified tree composed of the main arteries documented in the literature \cite{Assis15}.
This allowed us to reduce the problem to the annotation of a 17-branches binary tree that corresponds to the typical size of trees found in the literature \cite{Cao17}, \cite{Wang17}.
The selected arteries of interest are: the prostatic, the superior vesicle, the obturator (2), the pudendal, the inferior gluteals (2) and the superior gluteals (2).
Unlabeled branches in the tree correspond to proximal common portions of the arterial tree that we label "common" arteries for a total of 7 labels.
This simplified representation still captures the anatomical variability described in \cite{Assis15}, as we found 28 different tree labels arrangements among the 50 cases with high variability of branch shapes and positions as is illustrated in Fig.~\ref{fig:mean_shoot}.
Registrations are computed with the library \href{https://www.kernel-operations.io/keops/index.html}{KeOps} \cite{Keops2020} allowing fast GPU computing and automatic differentiation through \href{https://pytorch.org/}{PyTorch} (\cite{NEURIPS2019_9015}) backends. The optimization of the functional $\mathcal{J}$ is performed using Limited-memory Broyden–\-Fletcher–\-Goldfarb–\-Shanno (L-BFGS) algorithm.
In Fig.~\ref{fig:mean_shoot} (c) we plot the precision of the LDDMM-k at each iteration $k$. 

A first experiment was carried out to illustrate the contributions of the atlas construction on the labeling of the database. 
We computed LDDMM-0 (pure registration) and LDDMM-1 (atlas construction) by registering each of the 50 available cases onto the others.
It is illustrated in Fig.~\ref{fig:mean_shoot} (a,b).
The average precision of one reference case used in LDDMM-0 to annotate the 49 other trees is $93.3\%(\pm 3.5)$ when associated to bottom-up assignment and $84.2\%(\pm 4.4)$ using direct assignment.
This drop of performance illustrates the sensitivity of atlas-based methods to the choice of the atlas in the first place. 
The bottom-up assignment post processing allows to overcome this sensitivity: we use it in the rest of the experiments.
We then select one of the worst cases in the database regarding LDDMM-0 labeling performance and iteratively build the new atlas following the LDDMM-k procedure.

We can see that performance improves with iterations, which indicates that the atlas gradually captures the database variability: it allows a better registration hence a better label assignment.
This single-case solution allows to annotate the 49 cases of the database with a precision reaching $98.9\%(\pm 0.33)$ while being one of the worst with LDDMM-0.
It must be pointed out that the atlas construction did not rely on any other annotated case than the one initially selected.
We also computed LDDMM-1 using each tree as reference case and obtained an average score of $96.8\%(\pm 0.34)$, improving the average precision by $3.5\%$. 
In addition the standard deviation for LDDMM-1 confirms that iterating to build an atlas makes the LDDMM-k method robust to the initial choice of the atlas.
In Fig.~\ref{fig:mean_shoot} (c) we observe that one iteration of the atlas construction is enough to greatly improve the labeling of the entire database, then the performance slowly increases until iteration 4.
Therefore, and to limit the computational cost, we chose to use LDDMM-1 for the rest of the experiments.
\begin{figure}[t]
\minipage[t]{0.58\textwidth}
  \includegraphics[trim={0cm 0 0cm 0cm}, clip, width=1\textwidth]{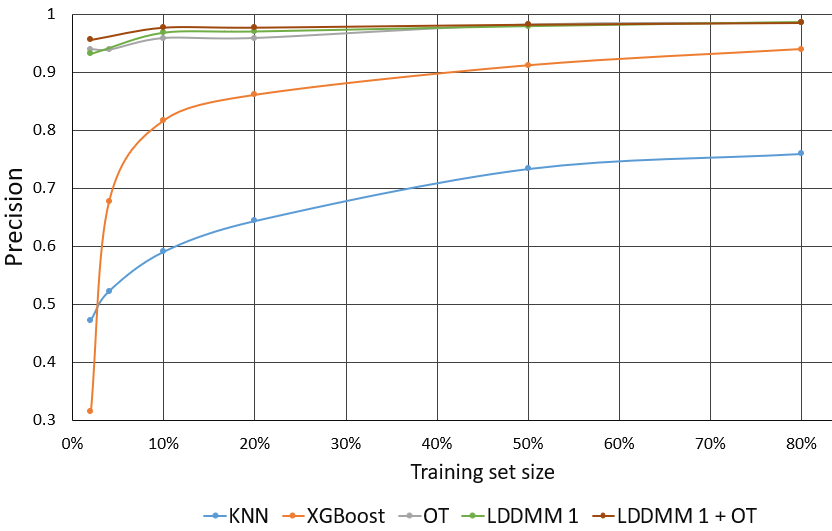}
  \captionsetup{labelformat=empty}
  \caption*{(a)}
\endminipage\hfill
\minipage[t]{0.4\textwidth}%
  \includegraphics[clip, width=1\textwidth]{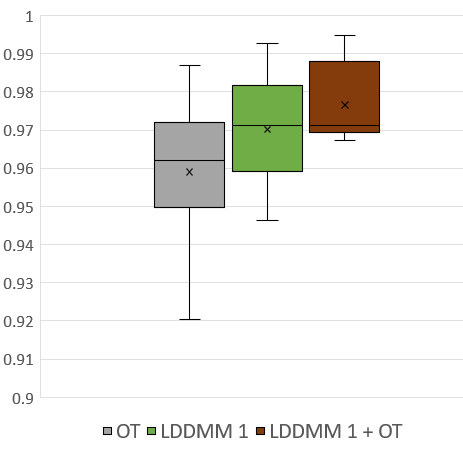}\\
  \captionsetup{labelformat=empty}
  \caption*{(b)}
\endminipage
\centering
\caption{\label{fig:ROC} (a) Comparison of the performance according to the training set size. (b) Box plot at training size 10\% (5 cases).}
\end{figure}

To demonstrate that atlas based techniques described in Sec.~\ref{sec:method} perform well compared to learning based ones in case of small size database, we implemented two classification algorithms working on branch features, KNN and XGboost, inspired from the work of \cite{Lo11} and \cite{Wang17}.
To be close to the work of \cite{Lo11} and \cite{Matsuzaki14}, each branch is represented by a vector composed of 28 \textit{branch features} such as direction, length or geometrical characteristics of the centerline curves,
and 13 \textit{tree features} involving branch's relationships to the root and to its children to introduce topological information.
These features are listed in the complementary material.
Tuning the XGBoost parameters had no impact so we kept parameters provided in \cite{Wang17}.
We performed 8-neigbours KNN in the space of the 10 most informative features selected by XGboost.

To compare all methods, precision has been evaluated using a cross-validation over the 50 cases with a training set and test set of varying size: from $2\%$ (1 case) to $80\%$ (40 cases) of the total dataset.
While the notion of cross-validation is well defined for learning-based techniques, in the case of atlas-based methods, it follows the leave-one-in procedure described in Sec.~\ref{Atlas} to select the best atlas among the training data.
This case is then used to annotate the test set.

For each method we used the assignment technique giving the highest precision (direct assignment for training-based methods and bottom-up for atlas-based ones). 
Results are presented in Fig.~\ref{fig:ROC}.
As expected, under 20\% of training data (10 cases) precision of learning-based methods drastically drops. KNN is outperformed by other approaches, and XGBoost seems to asymptotically reach the atlas-based performances. 
On the other hand the atlas-based methods with bottom-up assignment perform with very little influence of the size of the training set. 

Despite topological variations, the LDDMM-1 approach generates meaningful registrations showing good results when coupled with bottom-up assignment. 
OT also gives relevant branch matching that provides the same level of performance with the bottom-up assignment.
Consequently LDDMM-1+OT with the bottom-up assignment have the best results, particularly in the case of small training sets. 
We illustrate in Fig.~\ref{fig:ROC}~(b) the performances of atlas-based methods for only 5 cases in the training set (Confusion matrices for this setting are provided in the complementary material). 
The results of LDDMM-1+OT are significantly better than each method taken independently with an average $97.6\%(\pm0.97)$ precision.

\section{Conclusion and perspectives}

We have proposed an atlas-based labeling method allowing to annotate a database with a 97.6\% average precision using only 5 cases as training data. This level of precision isn't achieved by learning-based approaches even with 8 times more training data.
Our method takes advantage of the LDDMM realistic deformations to build a meaningful atlas and register it onto the cases to annotate. Optimal Transport computes a global branch matching that, combined with a bottom-up label assignment, provides an anatomically consistent labeling. 
The bottom-up assignment procedure allows to tackle the anatomical variations that usually degrades atlas-based methods precision. 
This procedure may be specific to pelvic anatomy, however we believe it could be easily adapted to other ones (\cite{Akinyemi09}, \cite{Assis15}, \cite{Lo11}). 
In addition, the LDDMM framework allows to perform statistics over deformations that could be exploited to generate new realistic data or to detect anomalies.
In further work we would like to extend our method to more complex cases such as missing labels or branches. In this perspective a promising lead would be to take topological changes into account as in \cite{Feragen15} for both the deformations and the data attachment.

\paragraph{Acknowledgement} We would like to thank Arthur Rocha, engineer at Sao Paulo Hospital das Clinicas for his help and expertise on the annotation of pelvic vascular trees.

{
\small
\bibliographystyle{splncs04}
\bibliography{MICCAI_bib}
}

\end{document}


\begin{table}
\centering
\label{Tab:BranchFeatures}
\begin{tabular}{|L{6cm}|c|}
\hline
Branch Features                                     & Dimension\\\hline \hline
Length                                              & 1        \\\hline
End-points distance                                 & 1        \\\hline
Main direction                                      & 3        \\\hline
End-points direction                                & 6        \\\hline
Average and standard deviation of the curvature     & 2        \\\hline
Invariant moments                                   & 3        \\\hline
Covariance matrix' eigen-values and -vectors        & 12       \\\hline
\end{tabular}
\caption{Branch features extracted for learning-based methods}
\end{table}

\begin{table}
\centering
\label{Tab:TreeFeatures}
\begin{tabular}{|L{6cm}|c|}
\hline
Tree Features per branch                          & Dimension\\\hline \hline
Number of descendant                              & 1        \\\hline
Ratio between average radius and parent branch    
average radius                                    & 1        \\\hline
Average and standard deviation of the average     
radius ratio between branch and children          & 2        \\\hline
Average and standard deviation of children        
root to end vectors                               & 6        \\\hline
Root-to-branch-barycentre vector                  & 3        \\\hline
\end{tabular}
\caption{Tree features extracted at the branch level for learning-based methods}
\end{table}

\begin{figure}
    
    \centering
    \mbox{
       \subfloat[\label{fig:Hungarian_from_leaves_confusion_matrix}]{%
          \includegraphics[clip, width=0.48\textwidth]{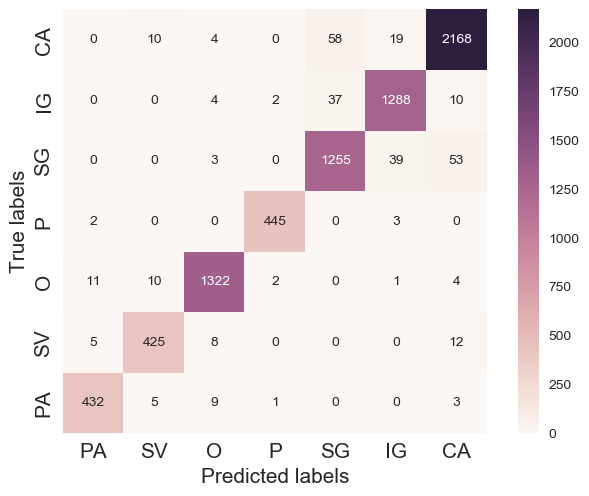}}
    \hfill
       \subfloat[\label{fig:confusion_matrix_LDDMM}]{%
          \includegraphics[clip, width=0.48\textwidth]{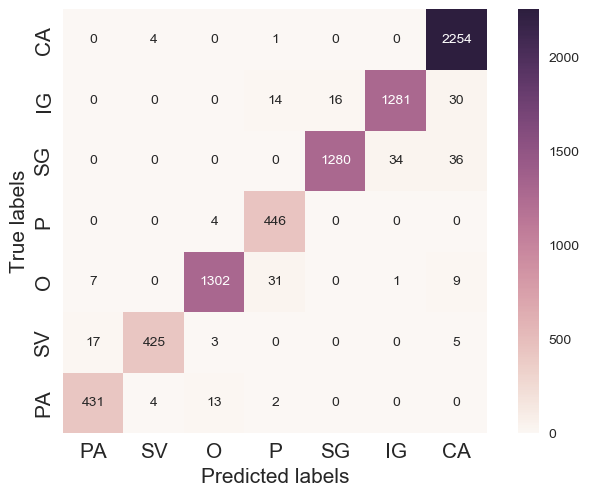}}
    }
    \\
    \mbox{
       \subfloat[\label{fig:confusion_matrix_LDDMM_OT}]{%
          \includegraphics[clip, width=0.48\textwidth]{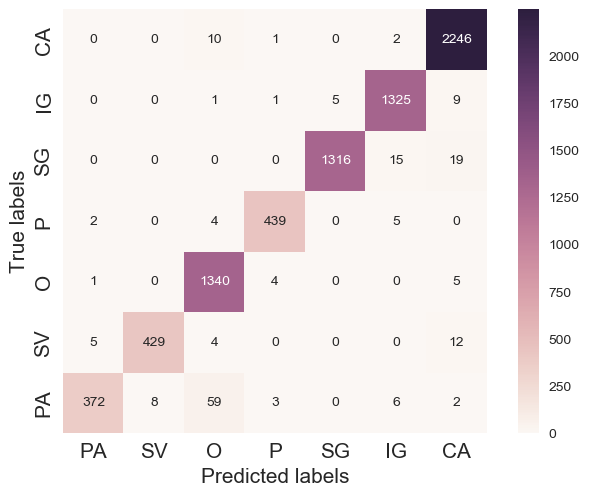}}
    \hfill
       \subfloat[\label{fig:XGB_from_leaves_confusion_matrix}]{%
          \includegraphics[clip, width=0.48\textwidth]{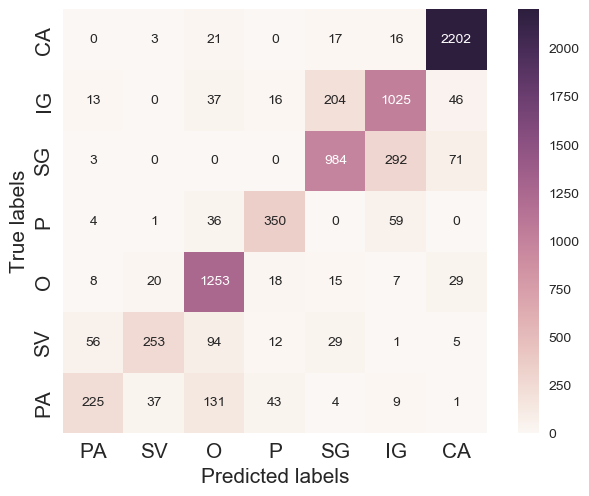}}
     }     
    \caption{\label{fig:confusion} 10-fold cross validation at training set size of 5 cases, label confusion matrices (a) OT (b) LDDMM 1 (c) LDDMM 1 + OT (d) XGBoost.}
    
\end{figure}